\documentclass[10pt,journal,compsoc]{IEEEtran}
\usepackage{pdfpages}
\usepackage{graphicx}
\usepackage{makecell}
\usepackage{mathrsfs}
\usepackage{array}
\usepackage{float}
\usepackage{amsmath}
\usepackage{hyperref}

\usepackage{epsfig}
\usepackage{epstopdf}
\usepackage{color}
\usepackage[british]{babel}
\usepackage{booktabs}
\usepackage{multirow}
\usepackage{url}

\ifCLASSOPTIONcompsoc
  \usepackage[nocompress]{cite}
\else
  \usepackage{cite}
\fi

\ifCLASSINFOpdf
 
\else
  
\fi

\usepackage{stfloats}

\fnbelowfloat

\begin{document}

\title{FCSR-GAN: Joint Face Completion and Super-resolution via Multi-task Learning}

\author{Jiancheng Cai, Hu Han*, ~\IEEEmembership{Member,~IEEE,} 
        Shiguang Shan, ~\IEEEmembership{Senior Member,~IEEE,} \\
        and Xilin Chen, ~\IEEEmembership{~Fellow,~IEEE}

\IEEEcompsocitemizethanks{\IEEEcompsocthanksitem J. Cai and X. Chen are with the Key Laboratory of Intelligent Information Processing, Institute of Computing Technology, Chinese Academy of Sciences, Beijing 100190, China, and also with the University of Chinese Academy of Sciences, Beijing 100049, China. E-mail: jiancheng.cai@vipl.ict.ac.cn; xlchen@ict.ac.cn.}

\IEEEcompsocitemizethanks{\IEEEcompsocthanksitem H. Han is with the Key Laboratory of Intelligent Information Processing of Chinese Academy of Sciences (CAS), Institute of Computing Technology, CAS, Beijing 100190, China, and also with Peng Cheng Laboratory, Shenzhen, China. E-mail: hanhu@ict.ac.cn.}

\IEEEcompsocitemizethanks{\IEEEcompsocthanksitem S. Shan is with the Key Laboratory of Intelligent Information Processing of Chinese Academy of Sciences (CAS), Institute of Computing Technology, CAS, Beijing 100190, China and University of Chinese Academy of Sciences, Beijing 100049, China, and he is also a member of CAS Center for Excellence in Brain Science and Intelligence Technology. E-mail: sgshan@ict.ac.cn.}

\IEEEcompsocitemizethanks{\IEEEcompsocthanksitem * Corresponding author: Hu Han.}

}

\markboth{IEEE TRANSACTIONS ON BIOMETRICS, BEHAVIOR, AND IDENTITY SCIENCE,~Vol.~\#\#, No.~\#\#, MM~2019}%
{Shell \MakeLowercase{\textit{et al.}}: Bare Advanced Demo of IEEEtran.cls for IEEE Biometrics Council Journals}

\IEEEtitleabstractindextext{
\begin{abstract}

Combined variations containing low-resolution and occlusion often present in face images in the wild, e.g., under the scenario of video surveillance. While most of the existing face image recovery approaches can handle only one type of variation per model, in this work, we propose a deep generative adversarial network (FCSR-GAN) for performing joint face completion and face super-resolution via multi-task learning. The generator of FCSR-GAN aims to recover a high-resolution face image without occlusion given an input low-resolution face image with occlusion. The discriminator of FCSR-GAN uses a set of carefully designed losses (an adversarial loss, a perceptual loss, a pixel loss, a smooth loss, a style loss, and a face prior loss) to assure the high quality of the recovered high-resolution face images without occlusion. The whole network of FCSR-GAN can be trained end-to-end using our two-stage training strategy. Experimental results on the public-domain CelebA and Helen databases show that the proposed approach outperforms the state-of-the-art methods in jointly performing face super-resolution (up to 8 $\times$) and face completion, and shows good generalization ability in cross-database testing. Our FCSR-GAN is also useful for improving face identification performance when there are low-resolution and occlusion in face images. The code of FCSR-GAN is available at: \url{https://github.com/swordcheng/FCSR-GAN}.
\end{abstract}

}

\maketitle

\IEEEdisplaynontitleabstractindextext

\IEEEpeerreviewmaketitle

\ifCLASSOPTIONcompsoc
\IEEEraisesectionheading{\section{Introduction}\label{sec:introduction}}
\else
\section{Introduction}
\label{sec:introduction}
\fi
\IEEEPARstart{C}{omplex} variations containing low-resolution and occlusions often exist in face images captured under unconstrained scenarios such as video surveillance. Obtaining high-resolution and non-occluded face images from low-resolution face images with occlusions is an essential but challenging task for face analysis such as face recognition, attribute learning, face parsing, etc.

While a number of approaches have been proposed for recovering high-quality face images from low-quality inputs, most methods aim at dealing with one type of variation per model, e.g., face completion \cite{paper1, song2018geometry}, and face super-resolution\cite{paper2, zhu2016deep, song2017learning, cao2017attention, chen2017fsrnet}. Under the assumption that there is only a single type of variation per image, the corresponding approaches for face super-solution or face completion can work well. However, these approaches may not fully meet the requirements of practical application scenarios where both low-resolution and occlusion may present simultaneously.

\begin{figure}[t]
  \centering
  \includegraphics[height=6.7cm]{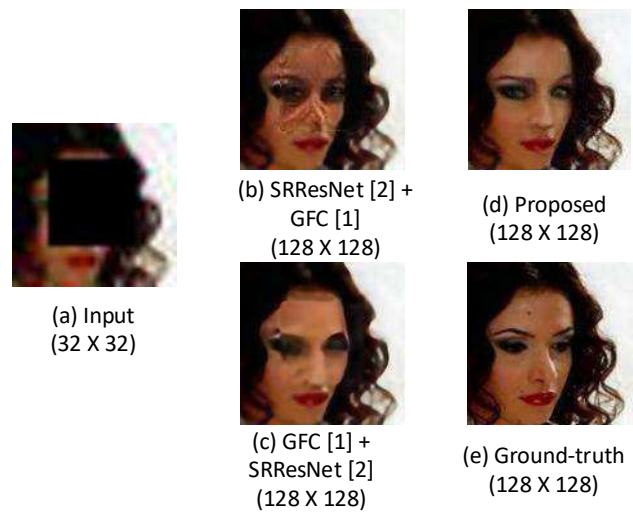}
  \caption{(a) Low-resolution input face image with occlusion (shown with 4$\times$ upsampling using the bicubic interpolation); (b) Recovered image by applying super-resolution (SRResNet \cite{paper2}) and face completion (GFC \cite{paper1}) successively; (c)  Recovered image by applying face completion (GFC \cite{paper1}) and super-resolution (SRResNet \cite{paper2}) successively; (d) Recovered image by our method; and (e) Ground-truth high-resolution face image without occlusion.}
  \label{fig:1}
  \vspace{-0.4cm}
\end{figure}

\begin{table*}[t]

\caption{A summary of recent representative methods on (face) image completion.} 
\centering 
\scalebox{1.25}{
\begin{tabular}{c c c c} 
\toprule[1pt]
\thead{Publication} & Method  & Dataset & Designed for face? \\ 
\toprule[1pt]
Pathak et al. \mbox{\cite{paper4_pathak2016context}}   & \thead{ CE \\ (Context Encoder-decoder structure)} &\thead{ Paris StreetView \cite{paris} \\ and ImageNet \cite{imagenet}} & No\\ 
Li et al. \mbox{\cite{paper1}} & \thead{ GFC \\ (Encoder-decoder structure; \\ global and local GAN)} & CelebA \mbox{ \cite{liu2015faceattributes} } & Yes\\
Yu et al. \mbox{ \cite{paper5_yu2018generative}} & \thead{ GntIpt \\ (Coarse-to-fine network architecture; \\  Contextual attention; \\global and local GAN)} &  \thead{Places2 \cite{place}, CelebA \cite{liu2015faceattributes}, \\ DTD textures \cite{dtd}, \\ and ImageNet \cite{imagenet}}  & No\\
Liu et al. \mbox{ \cite{liu2018image}} & \thead{ PConv \\ (Partial convolutional layer; \\ U-Net network architecture)} & \thead{Places2 \cite{place}, CelebA \cite{liu2015faceattributes}, \\ and ImageNet \cite{imagenet}}  &No\\

Zeng et al. \mbox{ \cite{zeng2019learning}} & \thead{ PEN-Net \\ (Pyramid filling; \\Cross-layer attention transfer)} & \thead{Facade \cite{tylevcek2013spatial}, DTD textures \cite{dtd}, \\ Places2 \cite{place}, CelebA \cite{liu2015faceattributes}}  &No\\

Sagong et al. \mbox{ \cite{sagong2019pepsi}} & \thead{ PEPSI \\ (Parallel extended-decoder; \\modified contextual attention)} & \thead{Places2 \cite{place}, CelebA \cite{liu2015faceattributes}, \\ and ImageNet \cite{imagenet}}  &No\\

Liu et al. \mbox{ \cite{liu2019coherent}} & \thead{ CSA \\ (coherent semantic attention; \\ Coarse-to-fine network architecture)} & \thead{Places2 \cite{place}, CelebA \cite{liu2015faceattributes}, \\ and Paris StreetView \cite{paris}}  &No\\

\bottomrule[1pt]
\end{tabular}
}
\label{tab1} 
\end{table*}

So different from existing approaches which mainly solve one challenge (either low-resolution or occlusion) per model, we aim at addressing a more challenging problem, i.e., how to handle both face low-resolution and occlusion in a single model. While a straightforward approach for handling both low-resolution and occlusion is to perform face super-resolution followed by face completion or vice versa, the effectiveness of existing face super-resolution approaches is not known when they are applied to low-resolution face image with occlusions\cite{paper2, zhu2016deep, song2017learning, cao2017attention, chen2017fsrnet}. Similarly, it is not known whether the face completion approaches work for low-resolution face images or not. As shown in Fig. \ref{fig:1}, when a face completion method (GFC \cite{paper1}) and a face super-resolution method (SRResnet \cite{paper2}) are applied successively to an input low-resolution face image with occlusion, the recovered face images (Fig. \ref{fig:1} (b) and (c)) may contain visual artifacts. The possible reason is that such a straightforward recovering approach is suboptimal because it treats super-resolution and de-occlusion as two independent problems; however, these two problems can have internal relationships during the image recovery process.  
In addition, when applying face completion and face super-resolution one after another, for Fig. \ref{fig:1} (b), artifacts may be introduced to the non-occluded region during the super-resolution, which then leads to more artifacts than the recovered face images in Fig. \ref{fig:1} (c).

\begin{table*}[t]
\caption{A summary of recent representative methods on (face) image super-resolution.} 
\centering 
\scalebox{1.08}{
\begin{tabular}{c c c c c} 

\toprule[1pt] 
Publication & Method  & Scale Factor & Dataset  & Designed for face?  \\
\midrule[1pt]
Dong et al. \mbox{\cite{dong2016image}} & \thead{SRCNN \\ (Three layers CNN; \\ Patch extraction and representation; \\ Non-linear mapping) }& $2\times, 3\times, 4\times$  & \thead{ImageNet \cite{imagenet}, \cite{Yang2010sr}, \\ Set5 \cite{set5}, Set14 \cite{set14}, \\ and BSD \cite{BSD}} & No \\ 
Kim et al. \mbox{\cite{kim2016deeply}} & \thead{DRCN \\ (Deeply-recursive convolutional structures) } & $2\times, 3\times, 4\times$ & \thead{ \cite{Yang2010sr}, Set5 \cite{set5}, \\ Set14 \cite{set14}, BSD \cite{BSD}, \\ and Urban \cite{urban}} & No \\
Ledig et al.  \mbox{\cite{paper2}} & \thead{SRResNet \\ (GAN structure; perceptual loss) } & $2\times, 4\times$ & \thead{ImageNet \cite{imagenet}, \\ Set5 \cite{set5},  Set14 \cite{set14}, \\ and BSD \cite{BSD}} & No \\
Zhu et al. \mbox{\cite{zhu2016deep}} & \thead{CBN \\ (Cascaded prediction; \\ bi-network architecture) } & $2\times, 3\times, 4\times$ & \thead{MultiPIE \cite{multipie}, \\ BioID \cite{bioid}, PubFig \cite{pubfig} \\ and Helen \cite{le2012interactive}}  & Yes \\
Cao et al. \mbox{\cite{cao2017attention}} & \thead{ AttentionFM \\ (Deep Reinforcement Learning; \\ Attended part enhancement) } & $4\times, 8\times$ & \thead{BioID \cite{bioid} \\ and LFW \cite{lfw}} & Yes\\
Song et al. \mbox{\cite{song2017learning}} & \thead{LCGE \\ (Face component generation \\ and enhancement) } & $4\times$ & \thead{Multi-PIE \cite{multipie} \\ and PubFig \cite{pubfig}} & Yes\\ 
Chen et al. \mbox{\cite{chen2017fsrnet}} & \thead{FSRNet \\ (Coarse-to-fine network architecture; \\ 
Using face landmark/parsing \\ maps prior knowledge)} & $8\times$ & \thead{CelebA \cite{liu2015faceattributes} \\ and Hellen \cite{le2012interactive}} & Yes\\ 

Guo et al. \mbox{\cite{guo2019adaptive}} & \thead{DCT-DSR \\(Convolutional discrete cosine transform; \\ orthogonally regularized; \\ image transform domain} & $2\times, 3\times, 4\times$ & \thead{Set5 \cite{set5},  Set14 \cite{set14}, \\ BSD \cite{BSD}, Urban \cite{urban}} & No\\ 

Dai et al. \mbox{\cite{dai2019second}} & \thead{SAN \\(Second-order attention; \\ non-locally enhanced residual group)} & $2\times, 3\times, 4\times, 8\times$ & \thead{Set5 \cite{set5},  Set14 \cite{set14}, \\ BSD \cite{BSD}, Urban \cite{urban}, \\ Manga109 \cite{manga} } & No\\ 

Li et al. \mbox{\cite{li2019feedback}} & \thead{SRFBN \\(Feedback mechanism; \\ curriculum-based training strategy)} & $2\times, 3\times, 4\times$ & \thead{Set5 \cite{set5},  Set14 \cite{set14}, \\ BSD \cite{BSD}, Urban \cite{urban}, \\ Manga109 \cite{manga} } & No\\ 
\bottomrule[1pt]
\end{tabular}
}
\label{tab2} 

\end{table*}

In this paper, we propose an end-to-end trainable framework based on a generative adversarial network (GAN) for joint face completion and super-resolution via a single model (namely FCSR-GAN). The generator of the FCSR-GAN performs face super-resolution and completion simultaneously, aiming for recovering a high-resolution face image without occlusion from an input low-resolution face image with occlusion. The discriminator of FCSR-GAN contains six losses: (i) an adversarial loss aiming for differentiating between real and generated face images; (ii) a pixel loss aiming for good reconstruction of non-occluded high-resolution face images; (iii) a perceptual loss aiming for obtaining photorealistic texture; (iv) a smooth loss aiming for penalizing color distortions along the boundaries of the occluded area; (v) a style loss aiming for maintaining the style between the recovered facial area and the non-occluded facial area, and (vi) a face prior loss aiming for obtaining a reasonable facial component topological structure. We also propose a two-stage training strategy that enables the network to be trained effectively end-to-end. The proposed approach is evaluated on the public-domain CelebA \cite{liu2015faceattributes}, and Helen \cite{le2012interactive} datasets and face images with natural low-resolution and occlusion.

The main contributions of this work include: (i) an efficient approach for jointly performing face super-resolution and completion with a single model via multi-task learning; (ii) a two-stage training strategy that enables the network to be effectively trained; and (iii) promising results compared with the baseline methods that applying the state-of-the-art face completion and face super-solution algorithms one after another. 

This work is an extension of our previous work of FG2019 \cite{jccaifg19}. The essential improvements over our previous work include: we have improved the loss design in order to improve the quality of the recovered face image; (ii) we have simplified the first-stage training of FCSR-GAN so that we can train it directly,  without dividing the first-state training into three steps as in \cite{jccaifg19}; (iii) we have shown the possibility of building a general framework that can leveraging existing face or image completion and super-resolution approaches to build an end-to-end recovering model; (iv) we have provided comprehensive review about related work and more details and evaluations of our FCSR-GAN.

\section{Related Work}

We briefly review the representative image completion and image super-resolution methods for either general images or face images. 

\subsection{Image Completion}

Image completion is to recover the missing content given an image with partial occlusion or corruption. 
Early image completion methods usually make use of the information of the surrounding pixels around the occluded region to recover the missing part. 
Ballester et al. \cite{ballester2001filling} proposed to perform joint interpolation of the image gray-levels and gradient directions to fill the corrupted regions. 
Such an approach may not work well when the corrupted region is large or has large variance in pixel values.  
Efros et al. \cite{paper3_image_inpainting} proposed a patch-based method to search relevant patches from the non-corrupted region of the image and use them to gradually fill the corrupted regions from outside to inside. 
While such an algorithm provides better results than previous methods, the patch search process can be slow.  
In order to solve this issue, Barnes et al. \cite{barnes2009patchmatch} proposed a fast patch search algorithm, but this method still cannot perform image completion in real-time. 
In general, the traditional methods usually rely on local context information but seldom consider the holistic context information in an image.
,
Recent efforts on image completion seek to utilize deep neural networks (DNNs) for image c,omp,letion. ,
The essence of this kind of method is to predict the missing part of the image by using all the information of the uncorrupted area. 
Pathak et al. \cite{paper4_pathak2016context} proposed an encoder-decoder structured network to perform image completion. 
Contextual attention mechanism and the surrounding features were used as reference to repair the corrupted image region  in \cite{paper5_yu2018generative, sagong2019pepsi, liu2019coherent}.
Liu et al. \cite{liu2018image} used the partial convolution neural network to gradually recover the missing pixels layer by layer. 
One advantage of their method is that it does not assume the missing image region must be of regular shapes, e.g., rectangle. While the above methods performed image completion at a single scale, \cite{zeng2019learning} proposed a pyramid-context encoder to use information of different scales to improve the image completion result.

Face completion differs from general image completion in that the structures and the shapes of different persons' faces are very similar, but the individual faces' textures are different from each other. 
Therefore, the face topological structure should be retained during face completion. 
Zhang et al. \cite{zhang2018demeshnet} proposed to perform face completion by moving meshy shelter on the face, which is effective for repairing a small area of corruption. 
To handle a large area of occlusion, Li et al. \cite{paper1} proposed a face completion GAN, in which a face parsing loss was introduced to maintain the face topological structure, and both global and local discriminators were used to ensure the quality of the completed face image. 
This approach reported promising results on the CelebA \cite{liu2015faceattributes} dataset; however, its effectiveness in repairing low-resolution face images with occlusion is not known.

We summary the recent representative image completion methods for in Table \ref{tab1} covering the method, datasets for evaluation, etc.

\subsection{Image Super-resolution}
 
Image super-resolution aims to recover a high-resolution image that retains the content but with more details from a low-resolution input image. 
In this paper, we focus on single image super-resolution, so multi-image or multi-frame based super-resolution approaches are not discussed here; we refer the interested readers to literature \cite{mf1, mf2}. 
There are two main categories of approaches for super-resolution from a single image. One category is edge enhancement based methods, e.g., through liner, bicubic or Lanczos \cite{duchon1979lanczos} filtering. 
These methods are very fast and do not require training, but the output high-resolution images often lack details. 
The other category is learning based methods, such as, patch-based \cite{pentland1991practical}, Markov Random Field (MRF) \cite{paper4_pathak2016context}, sparse representation \cite{adler2010shrinkage} and DNN \cite{dong2016image, paper2, kim2016deeply}  based methods. 
Benefiting from the strong modeling capacity of DNNs, the methods of \cite{dong2016image, paper2, kim2016deeply} perform much better than the traditional approaches. 
For example, Dong et al. \cite{dong2016image} proposed a SRCNN method for image super-resolution and reported much better results compared to the traditional methods. 
SRCNN is a lightweight network with a few layers and relatively small receptive field, so its fitting ability may be limited. 
Kim et al. \cite{kim2016deeply} proposed a DRCN with much deeper network than SRCNN. Still, recovering a high-resolution image with a large upscaling factor, e.g., 4$\times$, is found to be difficult. 
In order to get over this limitation, Ledig et al. \cite{paper2} proposed a perceptual loss for image super-resolution, which consists of an adversarial loss and a content loss. 
Guo et al. \cite{guo2019adaptive} proposed a DCT-DSR network to address the super-resolution problem in an image transform domain.
To further enhance the quality of the image, Dai et al. \cite{dai2019second} and Li et al. \cite{li2019feedback} proposed a second-order attention mechanism and a feedback mechanism to perform super-resolution, respectively.

\begin{figure*}[t]
  \centering
  \includegraphics[height=11.4cm]{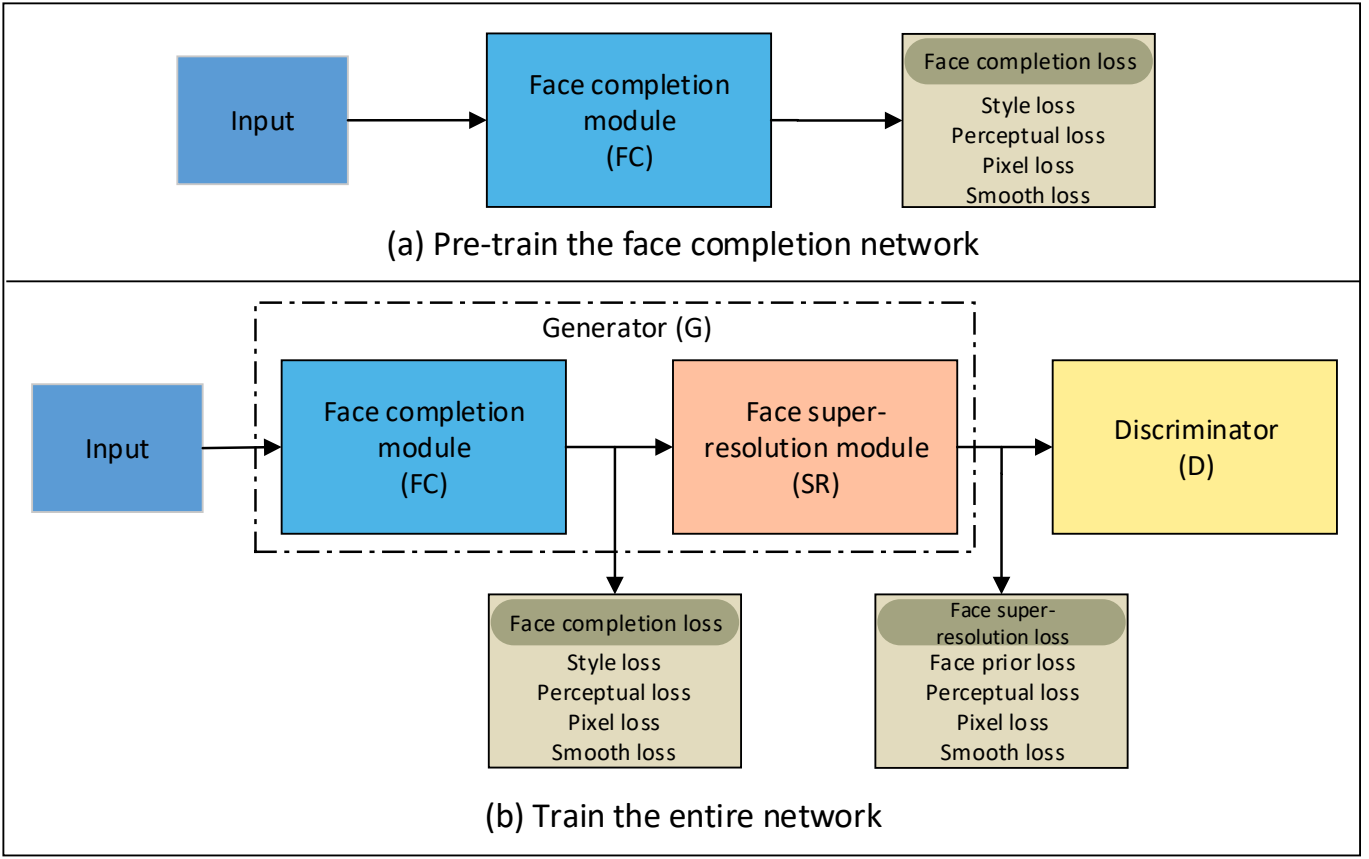}
  \caption{Two-stage training of the proposed FCSR-GAN for joint face image completion and super-resolution. (a) In the first stage, the face completion module is pre-trained using face completion loss; (b) In the second stage, the entire network is trained end-to-end by jointly using the adversarity loss, face completion loss, and face super-resolution loss.}
  \label{fig:2}
  \vspace{-0.2cm}
\end{figure*}

Face super-resolution is a special case of image super-resolution. 
Different from general image super-resolution, face super-resolution could make use of the domain knowledge of the face such as the face topological structure and the 3D shape information. 
Early face super-resolution methods were mainly motivated by general image super-resolution. 
Wang et al. \cite{wang2005hallucinating} utilized eigen transformation between the low-resolution space and high-resolution space to perform face super-resolution. 
This method assumes that the principal components of the low-resolution space and the high-resolution space are semantically aligned, but such an assumption may not hold in unconstrained scenarios where pose, illumination, and expression variations may exist. 
In addition, it could be difficult to perform face super-resolution with large-scale factors. 
Zhu et al. \cite{zhu2016deep} proposed a framework for hallucinating faces with unconstrained poses and very low resolution, in which framework, they alternatingly optimized two complementary tasks, namely face hallucination and dense correspondence field estimation. 
Cao et al. \cite{cao2017attention} made use of the attention mechanism and deep reinforcement learning to sequentially discover attended patches and then perform facial part enhancement by fully exploiting the global interdependency of the image. 
Song et al. \cite{song2017learning} and Chen et al. \cite{chen2017fsrnet} both used two-stage method to perform coarse-to-fine face super-resolution. The differences between them are that while \cite{song2017learning} first generated face components and then synthesized fine facial structures, but \cite{chen2017fsrnet} first generated coarse face images and then generated refined face images with more details.

We summary the recent representative image completion methods for general image and face image in Table \ref{tab2} covering the method, upscale factors, datasets in evaluation, etc.

\section{Proposed Method}

The overall framework of our FCSR-GAN is shown in Fig. \ref{fig:2} (b), which consists of a generator,  a discriminator, and the corresponding losses. As shown in Fig. \ref{fig:2}, we propose a two-stage training strategy for network learning. The face completion module is first trained as shown in Fig. \ref{fig:2} (a), and then the entire network is trained end-to-end as shown in Fig. \ref{fig:2} (b). We now provide the details below.

\subsection{Network Architecture}

\subsubsection{\textbf{Generator and Discriminator}}

\textbf{Generator.} As shown in Fig. \ref{fig:2} (b),  the generator of our approach is composed of a face completion (FC) module and a face super-resolution (SR) module, which aims to generate a non-occluded high-resolution image from a low-resolution face image with occlusion. 
Given an input low-resolution face image with occlusion $I_{LR}^{Occ}$, the output face image by the FC module is expected to be a non-occluded face image of the same size as the input. Ideally, only the occluded areas will be restored while the non-occluded areas should remain unchanged. The low-resolution face image after completion $I_{LR}$ by FC module can be represented as

\begin{equation}
    I'_{LR} = FC(I_{LR}^{Occ})
\end{equation}
\begin{equation}    
    I_{LR} = (1 - M) \odot I'_{LR} + M \odot I_{LR}^{Occ}, 
\end{equation}
where $\odot$, M and $I'_{LR}$ represent the pixel-wise dot product, the occlusion area (0 for occluded pixels and 1 for non-occluded pixels), and the direct output by FC module, respectively. 
Then, the face image $I_{LR}$ is input to the SR module to get high-resolution non-occluded face image. 
The final recovered high-resolution non-occluded face image can be computed as

\begin{equation}
    I_{HR} = SR(I_{LR}).
\end{equation}

We should point out that such a compound generator above can leverage the state-of-the-art image completion methods and super-resolution methods. Without losing generality, here, we choose to use either GFC \cite{paper1} or Pconv \cite{liu2018image} as the FC module, and use either SRResNet \cite{paper2} or FSRNet \cite{chen2017fsrnet} as the face SR module. Thus, we can have four different kinds of compound generators e.g., joint GFC and SRResNet, joint GFC and FSRNet, joint Pconv and SRResNet, joint Pconv and FSRNet. In our experiments, we can see that the proposed approach is effective for either of the above four kinds of generators. \emph{However, we should point out that joint face completion and super-resolution is not as simple as putting two methods together. We need to carefully design the loss functions and the training strategies so that we can leverage multi-task learning to obtain good face recovery results.} We will detail these in the following sections.

\textbf{Discriminator.} The discriminator $D$ of our approach plays an auxiliary role in network training. We use the commonly used real vs. fake discriminator for discriminating between real and fake (synthesized) face images. The structure of our discriminator is same as Patch-GAN \cite{patchgan}. The loss function is defined as

\begin{equation}
\begin{aligned}
L^{HR}_{adv} &  = \min_{\mathcal{G}} \max_{\mathcal{D}}  \mathcal{E}_{x \sim p_{data}(x)}[log \mathcal{D} (x)]  \\ &  +   \mathcal{E}_{z \sim p_{z}(z)}[log (1 - \mathcal{D} (\mathcal{G}(z)))], \label{eq:adv}
\end{aligned}
\end{equation}
where $p_{data}(x)$ represents the distribution of real face images and $p_{z}(z)$ represents the distribution of occluded face images.  The discriminator is only used in the second-stage training as shown in Fig. \ref{fig:2} (b).

\subsubsection{\textbf{Loss functions}}

\textbf{Perceptual loss.} When we recover a high-resolution face image from a low-resolution face image with occlusion, only using pixel-level mean absolute error (MAE) loss may lead to an over-smoothed image lacking details. 
We expect that both the recovered high-resolution images and the ground-truth high-resolution images should be as similar as possible from the perspective of low-level pixel values, high-level abstract features, and overall concept and style. 
In order to achieve this goal, we use perceptual loss in addition to pixel loss. 
The perceptual loss is defined as
\begin{equation}
\begin{aligned}
  \begin{cases}
      L^{LR}_{per} \  = &\frac{1}{W_{i,j}H_{i,j}}\sum\limits_{n=0}^{N-1} \Vert \phi_{i, j}(I'_{LR}) - \phi_{i, j}(I_{LR}^{gt}) \Vert_1 + \\  &\frac{1}{W_{i,j}H_{i,j}} \sum\limits_{n=0}^{N-1} \Vert \phi_{i, j}(I_{LR}) - \phi_{i, j}(I_{LR}^{gt}) \Vert_1 \\
      \\
      L^{LR}_{per} \  = &\frac{1}{W_{i,j}H_{i,j}}\sum\limits_{n=0}^{N-1} \Vert \phi_{i, j}(I_{HR}) - \phi_{i, j}(I_{HR}^{gt}) \Vert_1, 
\end{cases}
\end{aligned}
\end{equation}
where $\phi$ is VGG-16 \cite{simonyan2014very} which is pre-trained on ImageNet \cite{imagenet}, $\phi_{i, j}$ represents the feature map of the j-th convolution layer before the i-th max pooling layer; $W_{i, j}$ and $H_{i,j}$ represent the dimensions of the feature map. 
The pixel loss is defined as
\begin{equation}
  \begin{cases}
      L^{HR}_{pixel} = \Vert I_{LR} - I_{LR}^{gt} \Vert_1 \\
      \\
      L^{HR}_{pixel} = \Vert I_{HR} - I_{HR}^{gt} \Vert_1,
  \end{cases}
\end{equation}
where $\Vert \cdot \Vert$ is the $L_1$ norm and $I_{LR}^{gt}$ and $I_{HR}^{gt}$ represent the ground-truth low-resolution face image without occlusion and the ground-truth high-resolution face image without occlusion, respectively.

\textbf{Style loss.} When we perform face image completion, we need to make the style of the completed area as similar as possible to the non-occluded area. Therefore, we introduce style loss \cite{liu2018image} into the FC module to reduce the artifacts along the boundaries between the recovered area and the non-occluded area. The style loss is used to perform an autocorrelation (Gram matrix) on each feature map to ensure the style unification of the recovered face part and the non-occluded face part. The style loss is defined as
\begin{equation}
\footnotesize
\begin{aligned}
L^{s1}_{style} = & \sum_{n=0}^{N-1} \Vert K_{n}(\phi_n(I'_{LR})^T \phi_n(I'_{LR}) -  \phi_n(I_{LR}^{gt})^T \phi_n(I_{LR}^{gt})) \Vert_1 + \\ &  \sum_{n=0}^{N-1} \Vert K_{n}(\phi_n(I_{LR})^T \phi_n(I_{LR}) - \phi_n(I_{LR}^{gt})^T \phi_n(I_{LR}^{gt})) \Vert_1, 
\end{aligned}
\end{equation}
where $K_n$ is a normalization factor $1/(C_n\cdot H_n\cdot W_n)$ for the $n$-$th$ VGG-16 layer; $C_n$, $H_n$ and $W_n$ are the number, height and width of the feature map, respectively. The style loss is used in the second stage.

\textbf{Smooth loss.} When we perform face image completion, the completed face image may contain subtle color distortions along the boundaries of the occluded area. We introduce a smooth loss to reduce such color distortions, which is defined as

\begin{equation}
\begin{aligned}
  \begin{cases}
  L^{LR}_{smooth} = & \sum\limits_{i=0}^{W} \sum\limits_{i=0}^{H} \Vert I_{LR}(i, j+1) -  I_{LR}(i, j) \Vert_1 + \\ 
                    & \sum\limits_{i=0}^{W} \sum\limits_{i=0}^{H}  \Vert I_{LR}(i + 1, j) -  I_{LR}(i, j) \Vert_1 \\
                    \\
  L^{HR}_{smooth} = & \sum\limits_{i=0}^{W} \sum\limits_{i=0}^{H} \Vert I_{HR}(i, j+1) -  I_{HR}(i, j) \Vert_1 + \\
                    & \sum\limits_{i=0}^{W} \sum\limits_{i=0}^{H} \Vert I_{HR}(i + 1, j) -  I_{HR}(i, j) \Vert_1,
  \end{cases}
\end{aligned}
\end{equation}
where $W$ and $H$ are the width and height of the recovered face image by generator $G$, respectively. 

\begin{figure*}[t]
  \centering
  \includegraphics[height=4.9cm]{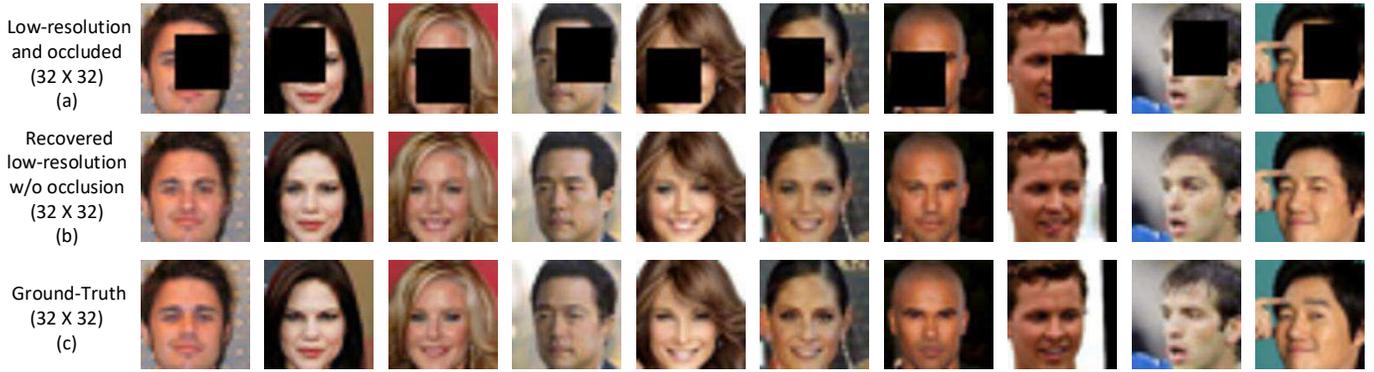}
  \caption{Face completion results by the face completion module of our FCSR-GAN from low-resolution face images with occlusions.  Since super-resolution module is not used, all the face images shown here are of low-resolution ($32\times32 $). For better visualization, each face images is shown with 4$\times$ upsampling using the bicubic interpolation.}
  \label{fig:low}
  \vspace{-0.3cm}
\end{figure*}
\begin{figure*}[t]
    \centering
    \includegraphics[height=7.5cm]{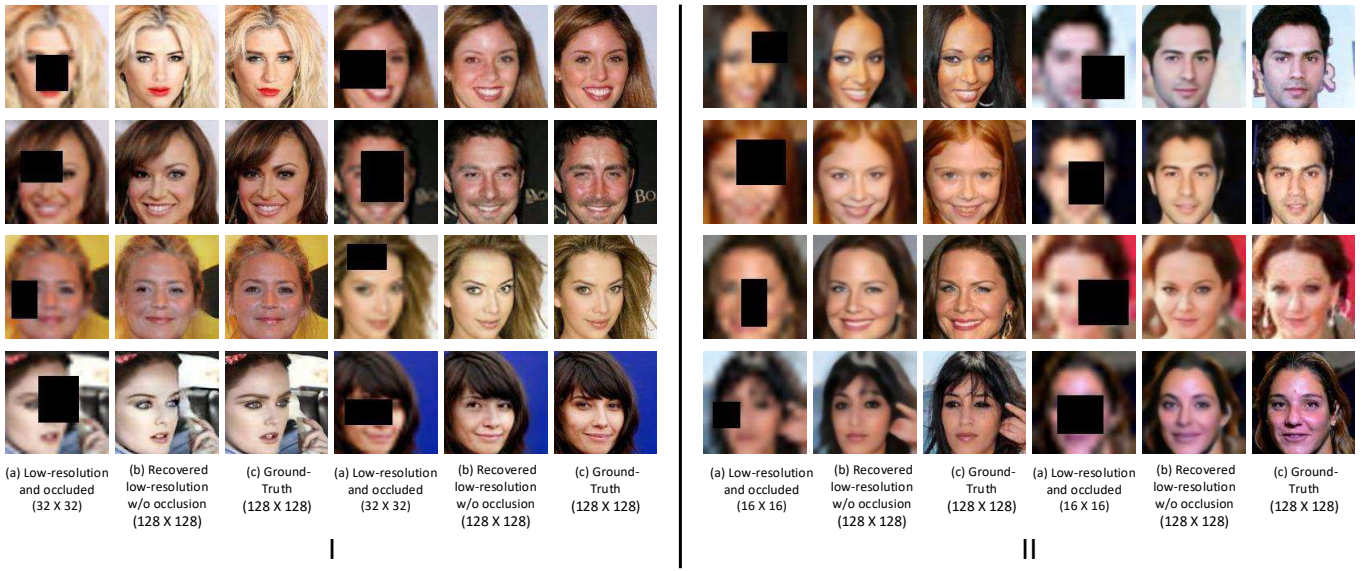}
    \caption{Joint face completion and super-resolution results by the proposed approach for some low-resolution face images with occlusions from CelebA. For panel I, (a) is the $32 \times 32$ input images (shown with 4$\times$ upsampling using the bicubic interpolation), (b) and (c) are the recovered and ground-truth face images. For panel II, (a) is the $16 \times 16$ input images (shown with 8$\times$ upsampling using the bicubic interpolation), (b) and (c) are the recovered and ground-truth face images.}
    \label{fig:different}
    \vspace{-0.3cm}
\end{figure*}

\textbf{Face prior loss.} The face contains several semantic parts (e.g., eyes, nose, mouth, etc.), which implies that a face has a more obvious topology than the other general objects. 
The discriminator above is to differentiate between real and generated face images from both global and local aspects so that the generator can generate more realistic face images. 
For example,  when the eye region of a face image is occluded, the generator is expected to recover the corresponding natural eye image without occlusion. 
However, in practice, the generator may generate face images with relatively strange facial geometry. In order to solve this problem, we refer to the face analysis method in \cite{chen2017fsrnet} and use a face prior as an auxiliary discriminator besides the real vs. fake discriminator.
This network can predict the face landmark heatmaps and face parsing maps simultaneously. The auxiliary discriminator is defined to penalize the difference between the predicted results and the ground-truth results.
The face prior loss is defined as
\begin{equation}
\begin{aligned}
L^{HR}_{fp} & = \alpha \Vert I_{landmark\_p} - I_{landmark\_gt} \Vert_{2} \\ & + \beta \Vert I_{parsing\_p} - I_{parsing\_gt} \Vert_{2}, 
\end{aligned}
\end{equation}
where $I_{landmark\_p}$, $I_{landmark\_gt}$, $I_{parsing\_p}$ and $I_{parsing\_gt}$ represent the predicted face landmark heatmaps from the generated face image, the ground-truth face landmark heatmaps, the predicted face parsing maps from the generated face image and the ground-truth face parsing maps, respectively. The face prior prediction network is used in the second stage.

\subsection{Two-stage Network Training}  \label{networktraining}

The end-to-end training of our generator is not as simple as training two separate modules independently. 
We design an effective two-stage training for the entire network as shown in Fig. \ref{fig:2}. As shown in Fig. \ref{fig:2} (a), the first-stage training uses style loss, perceptual loss, pixel loss, and smooth loss. In our experiments, we first pre-train FC (see Fig. \ref{fig:2} (a)). The entire loss at the first stage can be computed as
\begin{equation}
\begin{aligned}
    L_{fc} = & \lambda_{1}^{1} L^{LR}_{style} + \lambda_{1}^{2} L^{LR}_{per} \\ + & \lambda_{1}^{3} L^{LR}_{pixel} + \lambda_{1}^{4} L^{LR}_{smooth}, \label{eq:fc}
\end{aligned}
\end{equation}
where $\lambda_{1}^{1}$, $\lambda_{1}^{2}$, $\lambda_{1}^{3}$ and $\lambda_{1}^{4}$ are hyper-parameters balancing the influences of individual losses.  For the style loss, perceptual loss, pixel loss and smooth loss, we follow \cite{liu2018image} and use $\lambda_{1}^{1} = 10$, $\lambda_{1}^{2} = 0.1$, $\lambda_{1}^{3} = 0.1$, and $\lambda_{1}^{4} = 1$.

In the second training stage, we fix the FC module and use an adversarial loss, a face prior loss, a perceptual loss, a pixel loss, and a smooth loss to train the face super-resolution module.
As shown in Fig. \ref{fig:2} (b), the entire loss can be computed as
\begin{equation}
\begin{aligned}
L_{sr} =& \lambda_{2}^{1} L^{HR}_{adv} + \lambda_{2}^{2} L^{HR}_{fp} + \lambda_{2}^{3} L^{HR}_{per} \\ + & \lambda_{2}^{4} L^{HR}_{pixel} + \lambda_{2}^{5} L^{HR}_{smooth}, \label{eq:sr}
\end{aligned}
\end{equation}
where $\lambda_{2}^{1}$, $\lambda_{2}^{2}$, $\lambda_{2}^{3}$, $\lambda_{2}^{4}$, and $\lambda_{2}^{5}$ are hyper-parameters balancing different loss functions. Similarly, for adversarial loss and perceptual loss, we follow \cite{paper2} and use $\lambda_{2}^{1} = 10^{-3}$, $\lambda_{2}^{3} = 0.01$;  for face parsing loss, we follow \cite{chen2017fsrnet}, and use $\lambda_{2}^{2} = 1$; for the other losses, we empirically set $\lambda_{2}^{4} = 1$, $\lambda_{2}^{5} = 0.01$. 
{$L_{fc}$ and $L_{sr}$ are used to balance the loss items so that individual losses can be of the same order of magnitude, making it easier for the network to get convergence. }
Then, in the second training stage, we jointly train the whole network (see Fig. \ref{fig:2} (b)) using the  entire loss
\begin{equation}
L_{total} = L_{fc} + L_{sr}.
\end{equation}
Our ablation study in terms of the training strategy shows that such a two-stage training scheme can lead to better network convergence.

\begin{figure}[t]
  \centering
  \includegraphics[height=4.5cm]{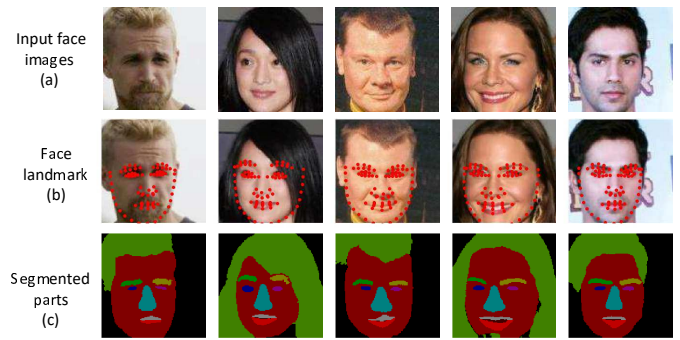}
  \caption{Face parsing results by the proposed approach on the CelebA dataset.   (a) is the input face images, (b) is the corresponding face  landmark localization results, and (c) is the corresponding face parsing results.}
  \label{fig:faceparsing}
  \vspace{-0.3cm}
\end{figure}

\begin{figure}[t]
    \centering
    \includegraphics[height=12cm]{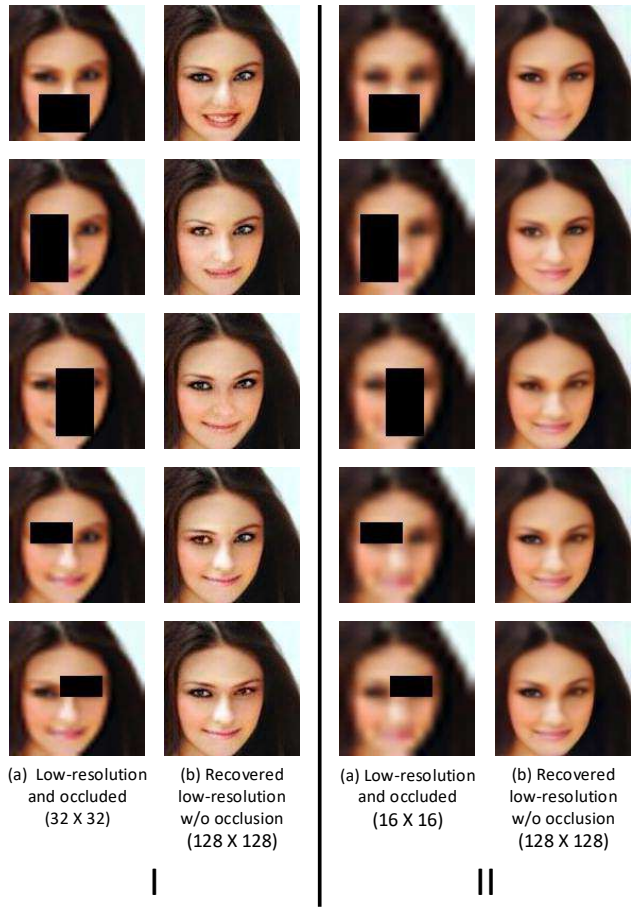}
    \caption{Joint face completion and super-resolution results by the proposed approach for face images of the same subject but with random occlusions at different locations of the face. For panel I, (a) is the input images (shown with 4$\times$ upsampling using the bicubic interpolation), and (b) is the recovered face images. For panel II, (a) is the input images (shown with 8$\times$ upsampling using the bicubic interpolation), and (b) is the recovered face images.}
    \label{fig:same}
\end{figure}

\begin{figure}[t]
  \centering

  \includegraphics[height=3.5cm]{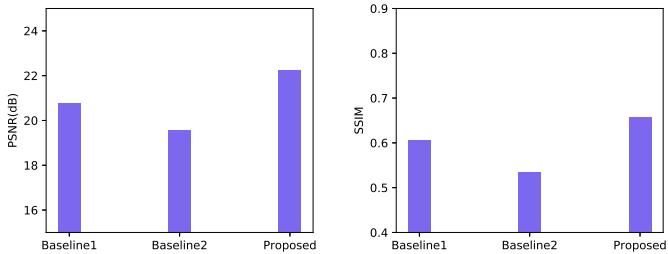}
  \caption{Quantitative comparisons between our FCSR-GAN and baselines in terms of PSNR and MSSIM under six types of occlusions. \textbf{Baseline1}: Results by applying face completion (GFC \cite{paper1}), and super-resolution (SRResnet \cite{paper2}) successively; \textbf{Baseline2}: results by applying face super-resolution (SRResnet \cite{paper2}) and face completion (GFC \cite{paper1}) successively; \textbf{Proposed}: our end-to-end trainable FCSR-GAN.}
  \label{fig:contract}
\end{figure}

\section{Experiment}

\subsection{Datasets}

We perform experimental evaluations on two public-domain datasets: CelebA \cite{liu2015faceattributes} and Helen \cite{le2012interactive}. CelebA is a large-scale face attribute dataset with 10,177 subjects and 202,599 face images; each is annotated with 40 binary attributes. We follow the standard protocol and divide the dataset into a training set (162,770 images), a validation set (19,867 images), and a test set (19,962 images). Helen is composed of 2,330 face images, and each image has 11 labels, denoting the main face parts such as eye, nose, mouth, etc. We follow the standard protocol of Helen and use 2,000 images for training and 330 images for testing. In our experiments of face completion and super-resolution,  the CelebA is used to train and test, and Helen is used to train the face parsing module of FCSR-GAN.

\subsection{Training Details}

We first train a face parsing network on Helen dataset.
We crop and align the face images from the Helen dataset based on the positions of the two eyes provided in the dataset. During training, the face images are aligned to $144\times144$ and then randomly cropped to $128\times128$ as inputs.
Random crop is a commonly used data augmentation method during network training, and is helpful to improve the network robustness. 
We use Adam \cite{kingma2014adam} algorithm with an initial learning rate $10^{-4}$ to optimize the face parsing network. 
Then, we use the trained face parsing network to predict the face parsing maps (see Fig. \ref{fig:faceparsing} (c)) for each face image in CelebA, and use these results as the ground-truth face parsing maps of CelebA. In addition, we also use an open-source SeetaFace \footnote{\url{https://github.com/seetaface/SeetaFaceEngine.}}
to locate 81 facial landmarks (see Fig. \ref{fig:faceparsing} (b)) for each face image in CelebA, and then using them as ground-truth facial landmarks of CelebA.

We  then perform two-stage training strategy for FCSR-GAN(described in Sect. \ref{networktraining}) from scratch using CelebA. We align and normalize each face image in CelebA to the same size of $128 \times 128$ following \cite{paper1}. 
In the first stage of training, we downsample each training image 4 times  (from $128 \times 128$ to $32 \times 32$) or 8 times (from $128 \times 128$ to $16 \times 16$) and introduce artificial occlusion to each face image (like the occlusions in Fig. \ref{fig:different}). 
In practice, it is very difficult to obtain paired face images, i.e., face images with natural occlusion and their mated face images without occlusion. Therefore, we use artificial occlusions to obtain paired face images. 
Some face completion results are shown in Fig. \ref{fig:low}.  In the second stage of training, the input image is the same as the first stage, but the output of the network is non-occluded high-resolution face image ($128 \times 128$).

\subsection{Experimental Results}

\subsubsection{\textbf{Qualitative comparisons}}

Qualitative comparisons can provide an intuitive observation of the recovered face images by different methods. 
We conducted two different experiments on the CelebA test dataset. 
Here, the face completion module we adopt is PConv \cite{liu2018image}, and the face super-resolution module is FSRNet \cite{chen2017fsrnet}. The input face images include both $\times 4$ or $\times 8$ times downsampled face images.
The first experiment is to verify the effectiveness of our FCSR-GAN in recovering a non-occluded high-resolution face image from a low-resolution image with occlusion at different locations (see Fig. \ref{fig:different}). 
We can see that the proposed approach can recover visually pleasing results compared with the ground-truth face images. 
The facial structures and important characteristics also look very similar to the ground-truth face images.
In some local details, the recovered face images are slightly different from the ground-truth. We think this is because face can have multiple perceptually reasonable states (e.g., open or closed eyes under sunglasses) and such differences with the ground-truth are reasonable. 
The second experiment is to verify the consistency of the proposed approach in recovering low-resolution face images of the same subject but with different occlusions (see Fig. \ref{fig:same}). Overall, the recovered high-resolution face images without occlusion of the same subject show very consistent facial components and details across individual images. This is a good characteristic of a face completion and super-resolution approach.

\subsubsection{\textbf{Quantitative comparisons}}

In addition to visual quality, we have also used two measurements to quantify the effectiveness of the proposed approach for joint face completion and super-resolution. One is the peak signal-to-noise ratio (PSNR), which is widely used in image compression area to measure the fidelity of the reconstructed signal w.r.t. the ground-truth. 
The other is the mean structural similarity index (MSSIM) \cite{wang2004image}, which is a perceptual measure that considers not only image degradation as perceived change in structural information, but also several perceptual phenomena, including luminance and contrast. 
For our FCSR-GAN, we use GFC and SRResNet as the face completion and super-resolution modules, respectively. In terms of the baselines, since there is not known prior methods that are reported to jointly handle low-resolution and occlusion, we use two straightforward baselines: successively performing face completion followed by face super-resolution (i.e., GFC + SRResNet), or vice versa (i.e., SRResNet + GFC).
In other words, we expect to evaluate the advantages of joint face completion and super-resolution via multi-task learning  \cite{7961738, 8411215, 8578652, 8606226}  applying face completion and face super-resolution models successively. 
The PSNR and MSSIM of individual methods are shown in Fig. \ref{fig:contract}. We can see that our FCSR-GAN performs much better than the two baseline methods in terms of both PSNR and MSSIM. This suggests that our FCSR-GAN can leverage multi-task learning to build effective models for jointly handling low-resolution and occlusions.

\begin{figure}[t]
  \centering
  \includegraphics[height=3.3cm]{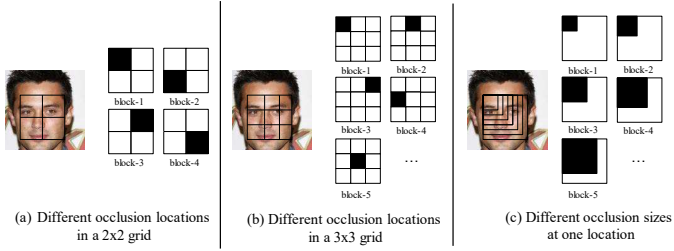}
  \caption{ Face completion and super-resolution by our FCSR-GAN with (a) and (b) different occlusion locations in low-resolution face images, and (c) different occlusions sizes. }
  \label{fig:occ_area}
\end{figure}

\begin{table}[t]
\caption{PSNR and MSSIM of face recovery by FCSR-GAN for the four different occlusion locations in Fig. \ref{fig:occ_area} (a).} 
\vspace{-0.1cm}
\centering 
\scalebox{1.0}{
\begin{tabular}{c c c c c c} 
\toprule[1pt] 
Block Index & Blk-1  & Blk-2 & Blk-3 & Blk-4 & Avg. and Std. \\ 
\midrule[1pt]
PSNR (dB) & 25.43 & 25.67 & 25.24 & 25.62 & 25.49 $\pm$ 0.197\\ 
MSSIM & 0.758 & 0.759  & 0.752 & 0.760 & 0.757 $\pm$ 0.0036\\ 
\bottomrule[1pt]
\end{tabular}
}
\label{tab:bl4} 
\end{table}

\begin{table*}[t]

\begin{center}

\caption{PSNR and MSSIM of face recovery by FCSR-GAN for the nine different occlusion locations in Fig. \ref{fig:occ_area} (b).}

\scalebox{1.08}{
\renewcommand{\arraystretch}{1.5}

\begin{tabular}{ m{1.4cm}<{\centering} m{1cm}<{\centering} m{1cm}<{\centering} m{1cm}<{\centering} m{1cm}<{\centering} m{1cm}<{\centering} m{1cm}<{\centering} m{1cm}<{\centering} m{1cm}<{\centering} m{1cm}<{\centering} m{1.2cm}<{\centering} }

\toprule[1pt] 
Occluded Blk. Index & Blk-1  & Blk-2 & Blk-3 & Blk-4 & Blk-5 & Blk-6 & Blk-7 & Blk-8 & Blk-9 & Avg. and Std.\\
\midrule[1pt]
PSNR (dB) & 27.67 & 27.41 & 27.40 & 27.10 & 27.24 & 27.11 & 27.62 & 27.43 & 27.29 & 27.36 $\pm$ 0.212\\

MSSIM & 0.778 & 0.780 & 0.789 & 0.764 & 0.768 & 0.769 & 0.775 & 0.778 & 0.776 & 0.775 $\pm$ 0.0075\\ 
\bottomrule[1pt]
\end{tabular}
}
\label{tab:bl9}
\end{center}
\vspace{-0.3cm}

\end{table*}

\begin{figure}[t]
  \centering
  \includegraphics[height=5.2cm]{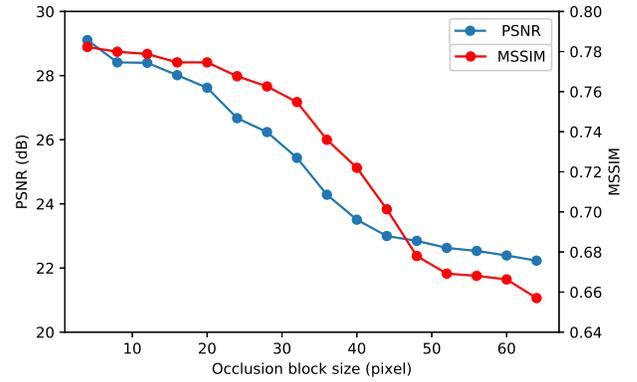}
  \caption{{Quantitative results (in PSNR and MSSIM) of joint face completion and super-resolution by our FCSR-GAN with different sizes of occlusion blocks.} }
  \label{fig:loc_size_com}
  \vspace{-0.3cm}
\end{figure}

In addition, we also study the influences of different occlusion sizes and positions to the face image recovery performance.
First, we divide the face images in the testing set into $2 \times 2$ grid (4 blocks in total as shown in Fig. \ref{fig:occ_area} (a)) and $3 \times 3$ grid (9 blocks in total as shown in Fig. \ref{fig:occ_area} (b)), and occlude one block each time to study the influences of different occlusion locations to our FCSR-GAN.  
The PSNR and MSSIM of FCSR-GAN at four different occlusion locations in Fig. \ref{fig:occ_area} (a) are given in Table \ref{tab:bl4}. Similarly, PSNR and MSSIM of FCSR-GAN at nine different occlusion locations in Fig. \ref{fig:occ_area} (b) are shown in Table \ref{tab:bl9}. From the results, we can see that (i) under the same occlusion block size, different occlusion locations have very small influence to the face image recovery performance (the standard deviations for PSNR and MSSIM across different locations are very small); (ii) occlusion blocks, e.g., block-1 and block-3 in Fig. \ref{fig:occ_area} (a), and block-4, block-5, and block-6 in Fig. \ref{fig:occ_area} (b), which are close to the eyes are more difficult for joint face completion and super-resolution (the PSNR and MSSIM are lower than the other occlusion locations); (iii) comparing the recovery results for Fig. \ref{fig:occ_area} (a) and Fig. \ref{fig:occ_area} (b), we can notice that bigger occlusion sizes are more challenging for FCSR-GAN. 
Then, we fix one occlusion location (shown in Fig. \ref{fig:occ_area} (c)) and change the occlusion block size to see its influence. 
The PSNR and MSSIM of FCSR-GAN is given in Fig. \ref{fig:loc_size_com}. 
Again, we notice that bigger occlusion sizes have bigger influence to the face recovery performance.

\begin{figure}[t]
  \centering
  \includegraphics[height=4.1cm]{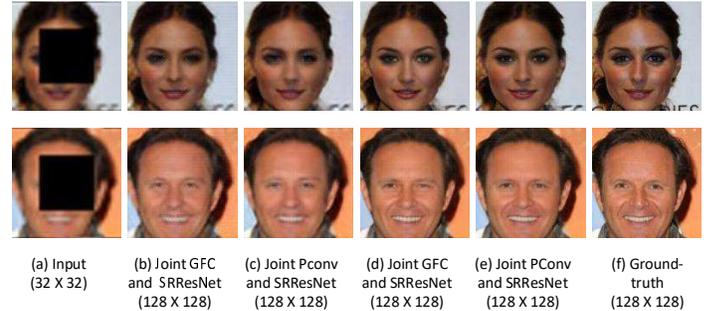}
  \caption{Recovered face images by our FCSR-GAN in {framework generality} test involving four different kinds of compound generators. (a) is the input low-resolution face image with occlusion. (b), (c), (d) and (e) are the results by joint GFC and SRResnet, joint PConv and SRResnet, joint GFC and FSRNet and joint PConv and FSRNet, respectively. (f) is the ground-truth high-resolution face image without occlusion.}
  \label{fig:12}
  \vspace{-0.3cm}
\end{figure}

\begin{table*}[t]
\tiny

\begin{center}

\caption{Ablation study of the proposed approach in terms of individual losses and the training strategy.}

\scalebox{1.6}{
\renewcommand{\arraystretch}{1.5}

\begin{tabular}{ m{1cm}<{\centering} m{1.1cm}<{\centering} m{1.1cm}<{\centering} m{1.1cm}<{\centering} m{1.1cm}<{\centering} m{1.1cm}<{\centering} m{1.3cm}<{\centering} m{1.3cm}<{\centering} m{1.1cm}<{\centering} }
\toprule[0.75pt] 
Ablation model & (a) Train together & (b) Stage 2 training alone & \makecell{(c) W/o \\ $L_{smooth}$} & (d) W/o $L_{p}$ & (e) W/o $L_{f}$ & \makecell{Proposed \\ FCSR-GAN}\\

\midrule[0.7pt]
PSNR (dB) & 20.18 &  20.09 & 22.14 & 21.72 & 21.94 & \textbf{22.23} \\
MSSIM & 0.573 & 0.595 & 0.652 & 0.626 & 0.649 & \textbf{0.657} \\
\bottomrule[0.75pt]
\end{tabular}
}
\label{tab3}
\end{center}
\vspace{-0.3cm}

\end{table*}

\subsubsection{\textbf{Ablation study}}

The proposed FCSR-GAN consists of several different loss functions, and uses a two-stage training scheme. In order to verify the effectiveness of each part, we perform five ablation experiments, i.e., (a) directly train the whole FSCR-GAN without using two-stage training;  (b) only perform the second-stage training  (without pre-training face completion as shown in Fig. \ref{fig:2} (a)); (c) train the model without smooth loss $L_{smooth}$; (d) train the model without perceptual loss $L_{per}$; and (e) train the model without face prior loss $L_{fp}$. Without losing generality, here, we use GFC as the face completion module and SRResNet as the face super-resolution module. All the experiments use the same input: 4 times downsampled image (i.e., from $128 \times 128$ to $32 \times 32$) with 1/4 area of occlusion. The results are shown in Table \ref{tab3}.  We can see that the proposed approach with two-stage training exceeds all the other ablation methods in terms of PSNR and MSSIM. The results suggest that each component of FCSR-GAN has its contribution for jointly face completion and super-resolution.

\subsubsection{\textbf{Framework generality}}

The proposed approach is actually a general framework in leveraging the state-of-the-art image completion methods and super-resolution methods to achieve joint face completion and super-resolution.
Without losing generality, here, we choose to use either GFC \cite{paper1} or Pconv \cite{liu2018image} as the FC module, and use either SRResNet \cite{paper2} or FSRNet \cite{chen2017fsrnet} as the SR module. 
Thus, we can have four different compound generators for our FCSR-GAN, e.g., joint GFC and SRResNet, joint GFC and FSRNet, joint Pconv and SRResNet, and joint Pconv and FSRNet. 
Fig. \ref{fig:12} shows some recovered face images by the above four  FCSR-GAN methods. 
All the experiments used the same input: 4 times downsampled image (i.e., from $128 \times 128$ to $32 \times 32$) with 1/4 area of occlusion. 
The PSNR and MSSIM of the compound generators methods are reported in Table \ref{tab:comp}. 
We can see that FCSR-GAN achieves better results when PConv and FSRNet are used as the compound generator, but the other three compound generators also work very well, indicating good generality of the proposed approach.

\begin{table}[t]

\caption{Quantitative results (in PSNR and MSSIM) by our FCSR-GAN in {framework generality} test involving four different kinds of compound generators.} 
\centering 
\scalebox{1.0}{
\begin{tabular}{c c c c} 
\toprule[1pt] 

Compound generator & Scale factor  & PSNR (dB) & MSSIM \\ 
\midrule[1pt]

Joint GFC and SRResNet & $\times 4$ & 22.23 & 0.657 \\ 
Joint PConv and SRResNet & $\times 4$ & 22.57  & 0.669\\ 
Joint GFC and FSRNet & $\times 4$ &  24.05&  0.749\\ 
Joint PConv and FSRNet & $\times 4$ &  \textbf{24.75}&  \textbf{0.761}\\ 
Joint GFC and FSRNet & $\times 8$ & 23.01 &  0.698 \\ 
Joint PConv and FSRNet & $\times 8$ & 23.77 & 0.710 \\ 
\bottomrule[1pt]
\end{tabular}
}
\label{tab:comp} 
\end{table}

\begin{figure}[t]
  \centering
  \includegraphics[height=4.5cm]{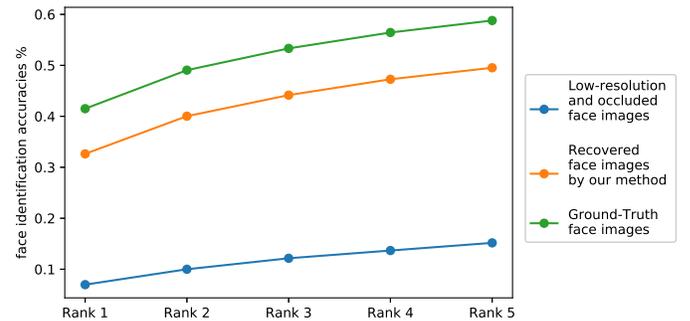}
  \caption{Face identification accuracies at rank 1-5 using low-resolution and occluded face images, recovered face images by our method, and the ground-truth face images, respectively.}
  \label{fig:face_recognition}
\end{figure}

\subsubsection{\textbf{Cross-dataset validation}}

We conduct cross-dataset validation to evaluate the generalization ability of our FCSR-GAN, i.e., training on CelebA but testing on Helen. Here, we use joint GFC and SRResNet as the compound generator. The input is $32 \times 32$ face images with 1/4 area of occlusion. Our FCSR-GAN trained on CelebA  achieves 21.72 dB PSNR and 0.628 MSSIM on Helen. Compared with the intra-database testing results on CelebA (22.23 dB PSNR and 0.657 MSSIM), these results look quite encouraging considering the different data distributions between CelebA and Helen.

\subsubsection{\textbf{Effectiveness for face recognition}}

We also explore whether the proposed FCSR-GAN can improving face recognition when using the recovered face images instead of the original low-resolution face images with occlusion in face recognition tasks. 
Here, we choose joint GFC and SRResNet as the compound generator. We conduct experiments on the CelebA dataset. The training set (containing 162,770 face images of 8,192 subjects) of CelebA used for face completion and super-resolution is also used as the training set for training face recognition model from a pretrained LightCNN-9 \cite{lightcnn}. For the testing set (containing 19,962 face images of 1,000 subjects) of CelebA, we randomly split it into gallery and probe, i.e., with one image of each subject in the gallery and the other images in probe. We used three types of face images for face recognition: (i) low-resolution and occluded face images, (ii) recovered face images by our FCSR-GAN, and (iii) the ground-truth high-resolution face images. In each face recognition experiment, the pre-trained LightCNN-9 is finetuned using the corresponding face images first, and then used for face identification. 
From the face identification results in Fig. \ref{fig:face_recognition}, we can see that face identification using the recovered face images by our FCSR-GAN leads to much higher accuracy than using the original low-resolution face images with occlusion, and the performance is close to that using the ground-truth high-resolution face images without occlusion. 
These results suggest that face completion and super-resolution by our approach is useful for improving face recognition performance.

\begin{figure}[t]
  \centering
  \includegraphics[height=6.5cm]{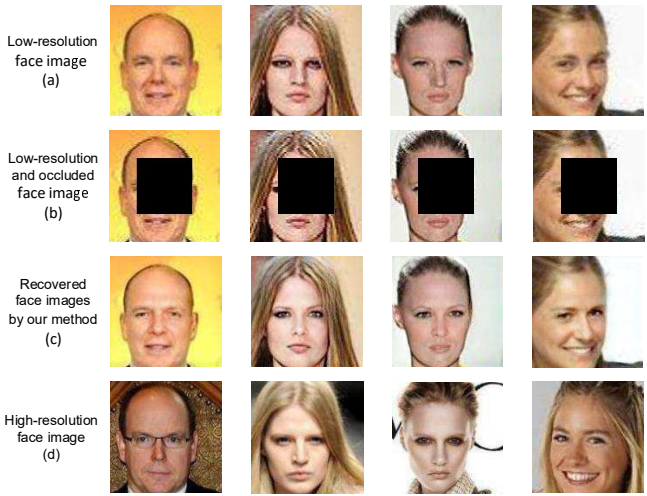}
  \caption{Face completion and super-resolution results by our approach for natural low-resolution and occluded face images. (a) low-resolution face image, (b) low-resolution face images with introduced occlusion, (c) the recovered high-resolution face images without occlusion by our FSCR-GAN, and (d) high-resolution face images from the same subject, which are used as references of the ground-truth.}
  \label{fig:natural_face_recover}
\end{figure}

\subsubsection{\textbf{Handling natural low-resolution face images}}

In the above experiments, we follow the state-of-the-art methods and use low-resolution face images that are downsampled from high-resolution face images. 
Downsampling can be different from natural low resolution in practice.
The main reason of using such a setting is that it is difficult to find paired face images with and without natural occlusion. 
We still expect to evaluate the effectiveness of our FCSR-GAN in handling natural low-resolution face images (see some examples in Fig. \ref{fig:natural_face_recover} (a)). 
For each natural low-resolution face image, we find a high-resolution face image of the same subject and use it as a reference of ground-truth (see Fig. \ref{fig:natural_face_recover} (d)).
For the natural low-resolution face images, we give the occlusion masks (see Fig. \ref{fig:natural_face_recover} (b)) and perform joint face completion and super-resolution using our FCSR-GAN. The recovered high-resolution face images without occlusion are shown in Fig. \ref{fig:natural_face_recover} (c). 
From the results, we can see that FCSR-GAN can recovering very reasonable face images compared with the reference ground-truth face images.

\begin{figure}[t]
  \centering
  \includegraphics[height=9cm]{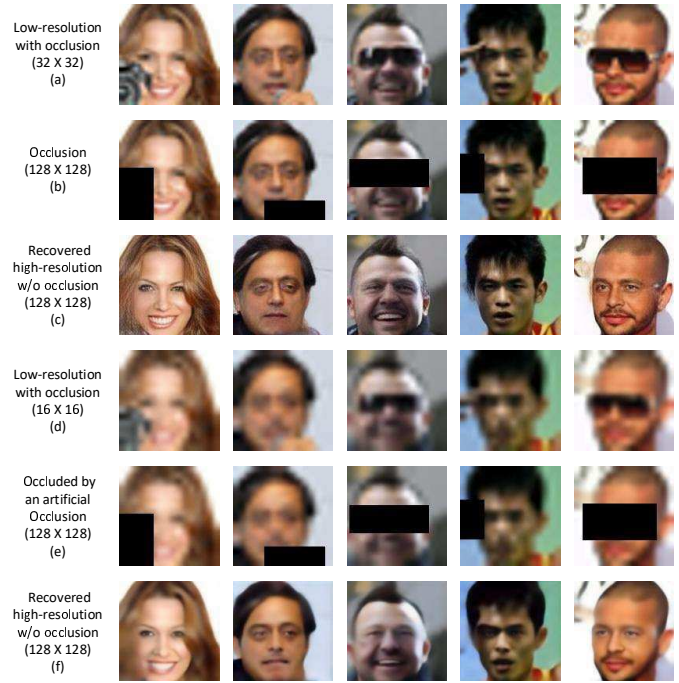}
  \caption{Handling low-resolution face images with natural occlusions in the wild. (a) and (d) are original low-resolution face images ($32 \times 32$ and $16 \times 16$, respectively) with natural occlusions by sunglasses, hand, etc.; (b) and (e) are input images ($32 \times 32$ and $16 \times 16$) to our FCSR-GAN network, in which the occlusion masks are provided; (c) and (f) are the recovered high-resolution ($128 \times 128$) face images without occlusion by our FSCR-GAN.}
  \label{fig:11}
\end{figure}

\subsubsection{\textbf{Handling natural face occlusion}}

The natural occlusion in practical applications (e.g., under video surveillance) can be different from the artificially generated occlusions like black blocks. 
We hope to evaluate the effectiveness of our FCSR-GAN in handling natural face occlusion. 
We find some naturally occluded low-resolution face images from CelebA, and give the occlusion masks. 
We then apply FCSR-GAN to recover the high-resolution face image without occlusion. 
We can see that although the ground-truth non-occluded high-resolution face images are not available, the recovered high-resolution face images without occlusion by FCSR-GAN look visual pleasing (see Fig. \ref{fig:11}). 
This suggests that the proposed FCSR-GAN might be used in some practical application scenarios.

\section{Conclusion}

In this paper, we propose a joint face completion and face super-resolution method (namely FCSR-GAN), which can leverage multi-task learning to recover non-occluded high-resolution face images from low-resolution face images with occlusions via a single model. 
The proposed FCSR-GAN uses compound generator and carefully designed losses (adversarial loss, perceptual loss, smooth loss, pixel loss, and face parsing loss) to assure the quality of the recovered face images. 
Experimental results on the public-domain CelebA and Helen databases show that the proposed approach outperforms the baseline methods in jointly performing face super-resolution (up to 8$\times$) and face completion from low-resolution face images with occlusions. 
The proposed approach introduces a general framework that can leverage the state-of-the-art image completion and super-resolution algorithms to achieve joint face completion and super-resolution. The proposed approach shows promising performance in both cross-dataset testing and in handling natural low-resolution and occlusion in face images.

Our current approach mainly deals with artificial occlusions, and requires input occlusion masks. 
In our future work, we would like to investigate methods for joint face completion and super-resolution from natural occlusions without providing manually labeled occlusion masks. 
In addition, we think the joint face completion and super-resolution problem is still far from being solved. 
For example, as discussed in \cite{pic}, face completion from occluded face images may have multiple reasonable de-occlusion results. 
It remains a challenging problem for balancing de-occlusion diversity and preserving subject identity. 
We also would like to study whether we can utilizing 3D face priors \cite{han20123d, niinuma2013automatic} to assist in the face completion and super-resolution task.

\ifCLASSOPTIONcompsoc
  \section*{Acknowledgments}
\else
  \section*{Acknowledgment}
\fi

This research was supported in part by the Natural Science Foundation of China (grants 61732004 and 61672496), External Cooperation Program of Chinese Academy of Sciences (CAS) (grant GJHZ1843), and Youth Innovation Promotion Association CAS (2018135).

\ifCLASSOPTIONcaptionsoff
  \newpage
\fi

\begin{IEEEbiography}[{\includegraphics[width=1in,height=1.25in,clip,keepaspectratio]{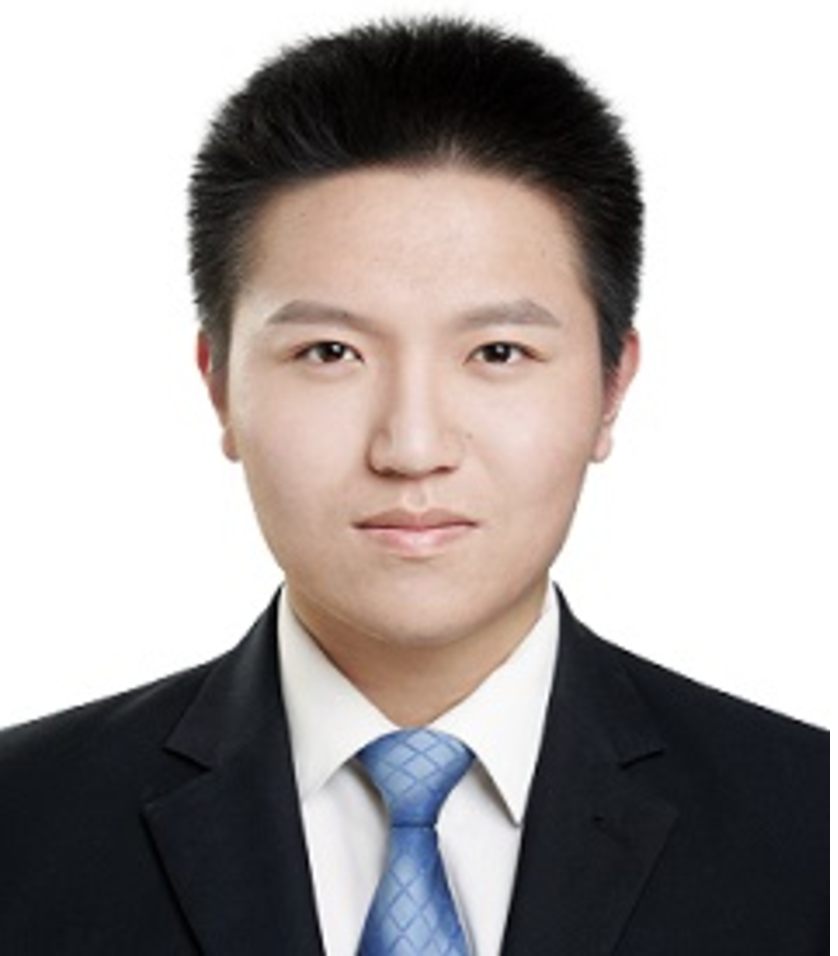}}]{Jiancheng Cai}
received the B.S. degree from Shandong University in 2017, and he is working toward the M.S. degree in the Institute of Computing Technology (ICT), Chinese Academy of Sciences (CAS), and the University of Chinese Academy of Sciences. His research interests include computer vision, pattern recognition, and image processing, with applications to biometrics.

\end{IEEEbiography}

\begin{IEEEbiography}[{\includegraphics[width=1in,height=1.25in,clip,keepaspectratio]{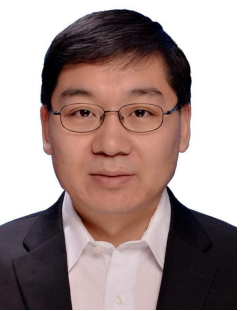}}]{Hu Han} 
is an Associate Professor of the Institute of Computing Technology (ICT), Chinese Academy of Sciences (CAS). He received the B.S. degree from Shandong University, and the Ph.D. degree from ICT, CAS, in 2005 and 2011, respectively, both in computer science. Before joining the faculty at ICT, CAS in 2015, he has been a Research Associate at PRIP lab in the Department of Computer Science and Engineering at Michigan State University, and a Visiting Researcher at Google in Mountain View. His research interests include computer vision, pattern recognition, and image processing, with applications to biometrics and medical image analysis. He has authored or co-authored over 50 papers in refereed journals and conferences including IEEE Trans. PAMI/IP/IFS, CVPR, ECCV, NeurIPS, and MICCAI. He was a recipient of the IEEE FG2019 Best Poster  Award, and CCBR 2016/2018 Best Student/Poster Awards. He is a member of the IEEE.

\end{IEEEbiography}

\begin{IEEEbiography}[{\includegraphics[width=1in,height=1.25in,clip,keepaspectratio]{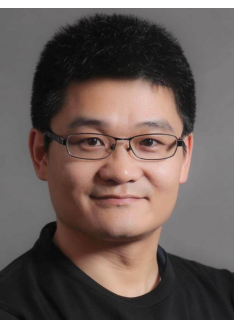}}]{Shiguang Shan}
is a Professor of ICT, CAS, and the Deputy Director with the Key Laboratory of Intelligent Information Processing, CAS. His research interests cover computer vision, pattern recognition, and machine learning. He has authored over 200 papers in refereed journals and proceedings in the areas of computer vision and pattern recognition. He was a recipient of the China¡¯s State Natural Science Award in 2015, and the China¡¯s State S\&T Progress Award in 2005 for his research work. He has served as the Area Chair for many international conferences, including ICCV11, ICPR12, ACCV12, FG13, ICPR14, and ACCV16. He is an Associate Editor of several journals, including the IEEE TRANSACTIONS ON IMAGE PROCESSING, the Computer Vision and Image Understanding, the Neurocomputing, and the Pattern Recognition Letters. He is a Senior Member of IEEE.

\end{IEEEbiography}

\begin{IEEEbiography}[{\includegraphics[width=1in,height=1.25in,clip,keepaspectratio]{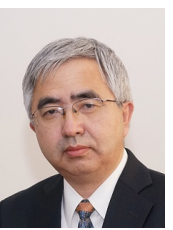}}]{Xilin Chen}
is a professor with the Institute of Computing Technology, Chinese Academy of Sciences (CAS). He has authored one book and more than 300 papers in refereed journals and proceedings in the areas of computer vision, pattern recognition, image processing, and multimodal interfaces. He is currently an associate editor of the IEEE Transactions on Multimedia, and a Senior Editor of the Journal of Visual Communication and Image Representation, a leading editor of the Journal of Computer Science and Technology, and an associate editor-in-chief of the Chinese Journal of Computers, and Chinese Journal of Pattern Recognition and Artificial Intelligence. He served as an Organizing Committee member for many conferences, including general co-chair of FG13 / FG18, program co-chair of ICMI 2010. He is / was an area chair of CVPR 2017 / 2019 / 2020, and ICCV 2019. He is a fellow of the IEEE, IAPR, and CCF.
\end{IEEEbiography}

\end{document}